\documentclass[a4paper]{article}

\usepackage{INTERSPEECH2022}
\usepackage{cite}
\usepackage{multirow}
\usepackage{amsfonts}
\usepackage{amsmath}
\usepackage{hyperref}

\title{Text-driven Emotional Style Control \\
and Cross-speaker Style Transfer in Neural TTS}
\name{Yookyung Shin, Younggun Lee, Suhee Jo, Yeongtae Hwang, Taesu Kim}
\address{
  Neosapience, Inc.
  }
\email{\{ykshin, yg, suheejo, ythwang, taesu\}@neosapience.com}

\begin{document}

\maketitle
\begin{abstract}
Expressive text-to-speech has shown improved performance in recent years. However, the style control of synthetic speech is often restricted to discrete emotion categories and requires training data recorded by the target speaker in the target style. In many practical situations, users may not have reference speech recorded in target emotion but still be interested in controlling speech style just by typing text description of desired emotional style. In this paper, we propose a text-based interface for emotional style control and cross-speaker style transfer in multi-speaker TTS. We propose the bi-modal style encoder which models the semantic relationship between text description embedding and speech style embedding with a pretrained language model. To further improve cross-speaker style transfer on disjoint, multi-style datasets, we propose the novel style loss. The experimental results show that our model can generate high-quality expressive speech even in unseen style.

\end{abstract}

\section{Introduction}

Recent advances in neural TTS systems enabled high-quality speech generation to be indistinguishable from natural speech \cite{shen2018natural, li2019neural, valle2020flowtron, ren2020fastspeech, kim2020glow}. 
Benefiting from this, neural TTS systems are widely used in real-world applications like AI voice assistants and navigation systems. Because synthetic speech that only sounds natural in a neutral style is insufficient for the expansion of the application of TTS to a variety of audio broadcasting scenarios like audiobooks and dubbing, there has been an increasing interest in expressive TTS and the method to control the synthetic speech style. 

Expressive TTS can be trained in both supervised and unsupervised manner. \cite{lee2017emotional} first attempted to control emotion with discrete emotion label, \cite{li2021controllable, lei2022msemotts} added emotion strength control, and took multi-scale nature of emotion into account. However, the control of speech emotion is still limited to several predefined emotion categories which makes it hard to reflect subtle variants of emotion. For example, \textit{sad} can be expressed in various ways ranging from \textit{crying out loud}, \textit{weeping}, to \textit{depressed}, but the emotion category \textit{sad} restricts the expression to the averaged style. Another way to control emotion uses reference speech. Global style tokens (GST) \cite{wang2018style} proposed a reference encoder that extracts style embedding from the acoustic signal. Reference speech-based expressive TTS can synthesize more diverse speech, but GST contains entangled representations of speaker, language, and style which makes it hard to interpret and intuitively control speech style. To alleviate this problem, \cite{kim2021expressive} first integrated a pretrained language model to control speaking style and emotion with style tags, which is the short description of speaking style written in natural language. 

Most expressive TTS models are trained on large single speaker datasets \cite{lee2017emotional, li2021controllable, lei2022msemotts, kim2021expressive}. Recording and annotating large-scale single-speaker expressive TTS data which covers a wide range of emotional styles is incredibly challenging. In reality, even the same sentence with the same emotion label can be expressed in various speech styles depending on the voice actor's interpretation and acting style. Or one person might be unable to record a variety of styles due to a lack of acting skill, but still want to synthesize speech in a specific style. An effective solution to this problem is to perform cross-speaker speech style transfer. As part of the solution, there have been researches on style transfer TTS with disjoint datasets, where a certain style is recorded by one speaker but the style is not present in other speakers \cite{whitehill2019multi, li2021controllable_cross, pan2021cross}. However, all existing methods are focused on emotion transfer using discrete emotion labels which ignores the complex nature of emotion conveyed in human speech.

In this paper, we propose a new approach to control and transfer emotional style in multi-speaker TTS. Our model can manipulate the emotional style of speech with text description by adopting the pretrained language model, which enables unlimited subtle control of emotion. Furthermore, because our model is trained on a disjoint, multi-style, and multi-speaker dataset, expressive emotional style can be learned and easily transferred between speakers. To improve the seen and unseen style transfer performance, we applied three additional style loss and data augmentation strategies. Experiments on seen and unseen emotional style transfer validated the effectiveness of our approach. 

\section{Related works}

\subsection{Language model in TTS}

Language model (LM) pretrained on large text corpora has shown successful results in many research areas. Since the typical neural TTS training requires speech-text paired data, it is more expensive to gather large-scale TTS corpora. There have been studies to improve TTS quality by leveraging the language understanding ability of pretrained LM like BERT \cite{devlin2018bert}. \cite{jia2021png} replaced the entire encoder of TTS system with BERT to learn better input text representation in order to improve prosody and pronunciation. In \cite{zhang2020unified}, BERT was applied in TTS front-end to improve G2P accuracy and to predict prosodic boundary. \cite{xu2021improving} modeled context information of neighboring sentences using BERT to improve prosody generation of synthesized speech in paragraph-level. BERT also has been integrated with TTS for style control purposes. \cite{lei2022msemotts} used BERT to predict emotion label from the text input, and \cite{kim2021expressive} to control speech style with style description written in text. Following \cite{kim2021expressive}, our model incorporated the pretrained BERT to learn the semantic relationship between text description embedding and speech style embedding.

\subsection{Cross-speaker style transfer}

Cross-speaker style transfer aims to synthesize speech in one speaker's voice with the reference style from another speaker. \cite{bian2019multi} introduced multi-reference encoder to \cite{wang2018style} and applied intercross training scheme, where each sub-encoder learns disentangled representation of style. In \cite{whitehill2019multi}, adversarial cycle-consistency training was proposed to ensure the use of all style combinations during training procedure via paired and unpaired triplets. \cite{sorin2020principal} used principal component analysis on style embedding, and \cite{pan2021cross} used prosody bottleneck, and \cite{li2021controllable} used emotion-speaker disentangling module with emotion classifier to disentangle style representation from text and speaker-related information. The proposed model differs from previous methods in that the control of style is not restricted to predefined emotion labels.

\section{Method}

\subsection{Style encoder}

Style encoder learns multi-modal embedding space by jointly optimizing a speech-side reference encoder and text-side style tag encoder with the acoustic model. 

\subsubsection{Reference encoder}

Reference encoder extracts speech-side style embedding from reference speech. It has almost the same architecture with mel-style encoder from Meta-StyleSpeech \cite{min2021meta}. Reference encoder is composed of 2 FC layers for spectral processing, 2 CNN layers with residual connection for temporal processing, followed by a multi-head self-attention with residual connection for gathering global information. The output from the final layer is temporally average pooled and used as a speech-side style vector \(\mathbf{w}_s\). In our experiments, this module has shown much better style reconstruction performance in multi-speaker training compared to the reference encoder of Global Style Tokens \cite{wang2018style}. 

\subsubsection{Style tag encoder}

Style tag encoder extracts text-side style embedding from style tag. We leveraged the generalization power of the language model pretrained on the massive amount of text as in \cite{kim2021expressive} in order to enable text-based manipulation of speech style, rather than learning the natural language understanding from scratch with a limited amount of text-to-speech data. We added an adaptation layer to map semantic embedding from SBERT \cite{reimers2019sentence} to style embedding space. The adaptation layer consists of three linear layers with ReLU activation. We used the pretrained Korean Sentence BERT (SBERT) model\footnote{https://github.com/BM-K/KoSentenceBERT-SKT}, and SBERT weights were fixed during the entire training.

\subsection{Acoustic model}

The backbone of the acoustic model is based on Tacotron2 \cite{shen2018natural}. As for the attention mechanism, we adopted the aligner and the duration predictor from \cite{badlani2021one}, which are more robust to long utterances and out-of-domain text, instead of the location-sensitive attention from Tacotron2. In consequence, the acoustic model consists of encoder, aligner, duration predictor, and decoder. 

\begin{figure}[t]
  \centering
  \includegraphics[width=\linewidth]{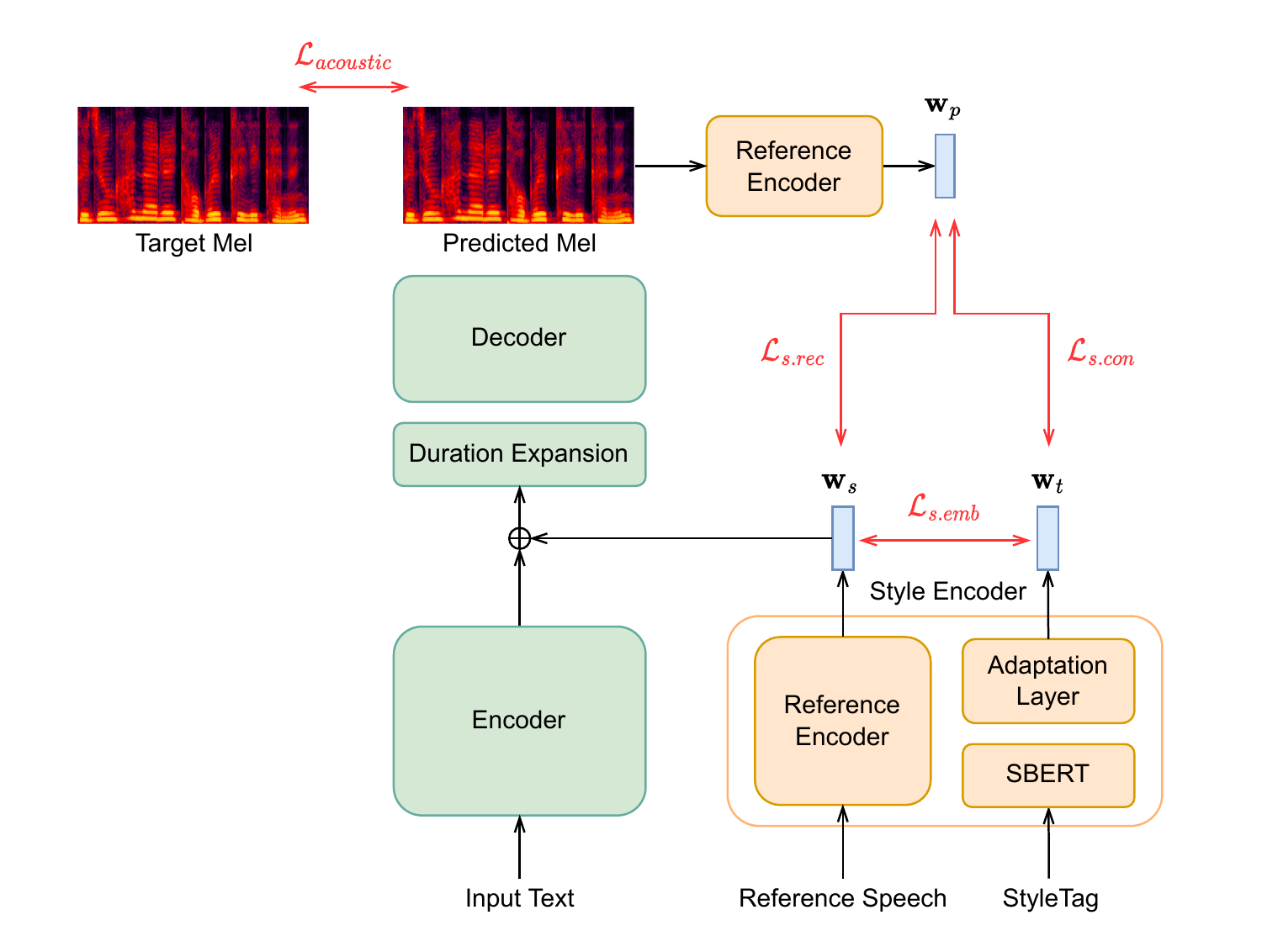}
  \caption{Model Architecture.}
  \label{fig:model}
\end{figure}

\subsubsection{Text encoder}

Text encoder takes the phoneme sequence as input and returns text embedding sequence as the output. It is composed of 3 CNN layers with batch normalization and ReLU activation followed by the BiLSTM layer. The encoder output is added with the style embedding then passed to aligner and duration predictor. This is because duration of each phoneme can differ depending on the speaking style. The encoder output sequence is expanded by the duration information from aligner for training and from duration predictor for inference and passed to decoder to match text and mel-spectrogram length.

\subsubsection{Aligner} \label{section:aligner}

Aligner aims to learn the alignment between mel-spectrogram and text. We followed the aligner module presented in \cite{badlani2021one}. Using two simple CNN encoders, phoneme sequence and mel-spectrogram are encoded into \(text^{enc}\) and \(mel^{enc}\) respectively. Then we compute the soft alignment based on the pairwise distance between all phonemes and mel frames which is then normalized with softmax across the text domain. Using the Viterbi algorithm, we then find the most likely monotonic path from the soft alignment to get hard alignment. Aligner is trained to maximize the likelihood of text given mel-spectrogram using the forward-sum algorithm. 
\begin{equation}
  D_{i,j} = dist_{L2}(text^{enc}, mel^{enc}),
  \label{eq1}
\end{equation}
\begin{equation}
  A_{soft} = softmax(-D, dim = 0).
  \label{eq2}
\end{equation}
\begin{equation}
  \mathcal{L}_{align} = \mathcal{L}_{ForwardSum} + (A_{hard} \odot log A_{soft})
  \label{eq3}
\end{equation}

\subsubsection{Duration predictor}

Duration predictor predicts the number of frames for each phoneme. Speaker embedding and style embedding are given as conditional input to model the style and speaker-dependent speaking speed. Duration predictor consists of 3 CNN layers with layer normalization. 

\subsubsection{Decoder}

Decoder is an autoregressive RNN which predicts mel-spectrogram from the encoder output as in \cite{shen2018natural}. The prediction from the previous time step first passes through prenet which consists of 2 fully connected layers with ReLU activation. The prenet output is concatenated with the context vector and the speaker vector and is passed through the LSTM layer followed by linear projection in order to predict target mel-spectrogram frame. The predicted mel-spectrogram finally passes through postnet which is consisted of 5 CNN layers aimed to predict the residual and added to the prediction to improve reconstruction quality. 

\subsection{Training and Inference} \label{section:loss}

The model is trained in an end-to-end manner on the sum of the following six loss terms. Along with conventional losses to train TTS system, we introduce novel objective functions to empower our model to control and transfer style.

First, following \cite{kim2021expressive}, style embedding loss is MSE loss between style embedding extracted from reference encoder \(\mathbf{w}_s\) and from style tag encoder \(\mathbf{w}_t\). Our model learns to map text-side style embedding and speech-side style embedding into shared multi-modal embedding space. Additionally, it provides supervision for the reference encoder to extract only emotion-related features, while discarding speaker-related features. 
\begin{equation}
    \mathcal{L}_{s.emb} = \sum_{i=1}^{N}(\mathbf{w}_{s_i}-\mathbf{w}_{t_i})^2
\end{equation}

Second, to enhance the emotion discriminative ability of predicted speech, we adopted style reconstruction loss, which is the MSE loss between style vectors extracted from predicted mel-spectrogram \(\mathbf{w}_p\) and reference mel-spectrogram \(\mathbf{w}_s\). Style tag and speech style do not have one-to-one mapping. A style tag by one speaker may be expressed differently by another speaker. When trained on style embedding loss alone, our model learned the averaged style of each style tag which resulted in degradation of expressiveness.
\begin{equation}
    \mathcal{L}_{s.rec} = \sum_{i=1}^{N}(\mathbf{w}_{s_i}-\mathbf{w}_{p_i})^2
\end{equation}

Finally, inspired by CLIP \cite{radford2021learning}, we adopted contrastive loss. It is essential that speech synthesized with similar style tags are located close to each other within the style embedding space. To learn such an embedding space, within a batch of \(N\) style embedding pairs, we maximized the cosine similarity of \(\mathbf{w}_t\) and \(\mathbf{w}_p\) for real pairs while minimizing the cosine similarity for negative pairs. We optimized a symmetric cross entropy loss over these similarity scores. This batch-wise contrastive representation learning technique has been widely utilized in speech, image, and text domains \cite{oord2018representation, zhang2020contrastive}. Without this loss, synthesized speech often expresses a different emotion from the desired one or fails to deliver consistent emotion within a sentence by showing an abrupt change of tone. We also observed that the speech quality was improved in extremely expressive samples such as screaming and crying.
\begin{equation}
    \mathcal{L}_\text{s.con} = - \log\frac{\exp(\text{sim}(\mathbf{w}_{t_i}, \mathbf{w}_{p_j} / \tau)}{\sum_{k \neq i, k=1,...,N} \exp(\text{sim}(\mathbf{w}_{t_i}, \mathbf{w}_{p_k}) / \tau)}
\end{equation}

The loss function to train the acoustic model is referred to as \(L_{acoustic}\). It includes \textit{mel reconstruction loss}, L1 loss between predicted mel-spectrogram and target mel-spectrogram, \textit{alignment loss}, the negative log likelihood loss for training aligner as described in \ref{section:aligner}, and \textit{duration prediction loss}, L2 loss between ground truth duration obtained by aligner and predicted duration. 
The final objective function is defined as: 
\begin{equation}
  \mathcal{L} = \mathcal{L}_{acoustic} + \alpha\mathcal{L}_{s.emb} + \beta\mathcal{L}_{s.rec} + \gamma\mathcal{L}_{s.con}
  \label{eq4}
\end{equation}
where \(\alpha = \beta = 1.0 \) and \(\gamma = 0.01\) are weights for different loss terms. 
During training, style embedding from reference encoder and style embedding from style tag encoder are alternatively chosen with random probability (\(p = 0.5\)) and are added to encoder output. When the model was trained with only style embedding from reference encoder, style transfer using style tag could not reach the expressiveness of the original data due to train-inference mismatch.  During inference, style control and transfer can be achieved either by style tag or reference speech. 

\section{Experiments}

\subsection{Data}

Our model is trained on the proprietary dataset which consists of 57 hours of Korean speech which were recorded by 38 professional voice actors. The amount of recording per speaker varies from 1 to 4 hours. To demonstrate seen and unseen style transfer performance of our proposed model, we construct a disjoint, multi-style dataset. Dataset can be divided into either source speaker data or target speaker data. Source speaker data is recorded in a neutral, reading-book style by 20 speakers. Target speaker data is recorded in a conversational style with 4 emotion categories (\textit{neutral}, \textit{happy}, \textit{angry}, and \textit{sadness}) by the remaining 18 speakers.

To enable text-guided control of speech style, crowd-sourced taggers were asked to label each input speech with appropriate style tags. Style tag is a short description of speaking style which includes informations regarding emotions, intentions, tone, and speed. Annotating preexisting TTS corpus with style tags is cheaper than recording the dataset from scratch, and because each speech can be annotated with multiple style tags, it ensures the diversity of style tags. As in FSNR0 \cite{kim2021expressive}, the final dataset consists of \{speech, transcript, style tag\} tuples. 

The total number of unique style tags is over 5000, and each style tag appears one to 4000 times. The number of style tags per speech varies from 1 to 8. During training, the random number of style tags are sampled and the mean of the selected style tags' embeddings are used as text style embedding for data augmentation. 

\subsection{Training setup}

Phoneme sequences are used as the input, which are obtained from text-normalization and grapheme-to-phoneme converter. The recordings are down-sampled from 48 kHz to 16kHz. We extracted a spectrogram with FFT size of 1024, window size of 50ms, and hop size of 12.5ms. Then, we converted it to a mel-spectrogram with 120 frequency bins. We used Hifi-GAN \cite{kong2020hifi} as the vocoder to convert mel-spectrogram into waveform. 

\subsection{Evaluation}
The goal of our model is to synthesize speech with desired emotional style guided either by style tag or reference audio with the voice of the target speaker. Therefore, we evaluated the performance of seen and unseen style transfer through subjective evaluation in terms of speech naturalness, style similarity, and speaker similarity. We conducted Mean Opinion Score (MOS) test for speech naturalness and Similarity MOS (SMOS) test for style and speaker similarity. Ten native Korean speakers are invited to participate in each experiment. Both metrics were rated in a 1-to-5 scale and reported with the 95\% confidence intervals. Audio samples can be found at the demo page \url{https://style-tts.github.io/demo/}.

\section{Results}

\subsection{Naturalness}

We conducted Mean Opinion Score (MOS) test for speech quality and naturalness. The results of the five groups: ground truth (GT), GT with vocoder reconstruction, Tacotron2-GST (baseline), and proposed model trained on a multi-speaker dataset are shown in Table~\ref{tab:t1_mos}. Proposed (reference) and proposed (style tag) mean the style of synthesized speech is controlled using reference speech and style tag respectively. \textit{Seen style transfer} refers to synthesizing speech in the target speaker's voice with emotion transferred from the target speaker itself. \textit{Unseen style transfer} refers to synthesizing speech in the source speaker's voice with emotion transferred from the expressive target speaker. 2 speakers (1 female, 1 male) were selected from neutral source speaker dataset for unseen style transfer, and 2 additional speakers (1 female, 1 male) were selected from emotional target speaker dataset for seen style transfer. 20 sentences were randomly selected from the test set to synthesize speech, and corresponding reference audios and style tags were used to transfer style. 

The result shows that the proposed model outperforms the baseline both in seen and unseen style transfer. Seen style transfer shows slightly better speech quality than unseen style transfer because the reference style has been seen during training. 

\begin{table}
    \caption{MOS Result on seen and unseen style transfer.}
    \label{tab:t1_mos}
    \centering
        \begin{tabular}{l|ccc}
            \noalign{\smallskip}\noalign{\smallskip}\hline\hline
            & \textbf{Seen} & \textbf{Unseen} \\
            \hline
             GT & 4.48 \(\pm\) 0.12 & - \\
             GT + Vocoder & 4.23 \(\pm\) 0.14 & - \\ 
            \hline
             Tacotron2-GST & 3.72 \(\pm\) 0.14 & 3.40 \(\pm\) 0.14 \\
            \hline
             Proposed(reference) & 4.13 \(\pm\) 0.15 & 3.77 \(\pm\) 0.19 \\
             Proposed(style tag) & 4.02 \(\pm\) 0.17 & 3.67 \(\pm\) 0.19 \\	
            \hline
            \hline
        \end{tabular}
\end{table}

\subsection{Style transfer}

We further conducted Similarity MOS (SMOS) test on style similarity to evaluate how well the transferred speech matches the emotional style of the reference and also on speaker similarity to evaluate how well the transferred speech matches the voice of the original speaker. For style transfer using reference speech, participants were given reference speech and transferred speech pair, and for style transfer using style tag, participants were given style tag and transferred speech pair. They were asked to evaluate how well two stimuli matches in terms of emotional style. In practical settings, the transcription of reference speech may be different from the input text, which is called non-parallel transfer. However, parallel transfer where input text matches that of reference speech is used for the evaluation task because it is easier to evaluate style similarity objectively if the text matches. Non-parallel transfer examples are presented on the demo page.

According to the results shown in Table~\ref{tab:t2_style_sim}, seen style transfer using reference speech successfully reconstructed the given style. This shows our reference encoder module can extract emotion-related features from reference audio. Unseen style transfer using reference speech showed degraded performance compared to the seen-style transfer due to the entangled speaker-style information in style embedding. The gap between seen and unseen style similarity scores is reduced when using style tag in style transfer. This means that unseen style transfer successfully generated emotional speech matching the input text description even though it could not reach the expressiveness of the source data. This validates the effectiveness of the proposed style contrastive loss as described in \ref{section:loss}. Without this loss, even the seen-style transfer failed to synthesize emotional speech matching the input style tag. 

Unseen style transfer showed higher speaker similarity compared to seen style transfer in Table~\ref{tab:t3_spkr_sim}. Because the degree of expressiveness in unseen style transfer was slightly lower, speaker identity was better preserved in synthesize speech. We reviewed test samples with the lowest speaker similarity score and found out that reference and synthesized speech have a big difference in emotion (\textit{e.g.} happy reference speech expressed in high pitch presented with sad synthesized voice expressed in low pitch), making it difficult for raters to recognize the speaker conformity.

\begin{table}
	\caption{Style similarity MOS result.}
	\label{tab:t2_style_sim}
	\centering
	\begin{tabular}{l|cc}
	\noalign{\smallskip}\noalign{\smallskip}\hline\hline
	& \textbf{Seen} & \textbf{Unseen} \\
	& emotional-to-emotional & neutral-to-emotional \\
	\hline
	Reference & 4.10 \(\pm\) 0.13 & 3.09 \(\pm\) 0.16 \\
    Styletag & 3.83 \(\pm\) 0.16 & 3.53 \(\pm\) 0.16 \\
	\hline
	\hline
	\end{tabular}
\end{table}

\begin{table}
	\caption{Speaker similarity MOS result.}
	\label{tab:t3_spkr_sim}
	\centering
	\begin{tabular}{l|cc}
	\noalign{\smallskip}\noalign{\smallskip}\hline\hline
	& \textbf{Seen} & \textbf{Unseen} \\
	& emotional-to-emotional & neutral-to-emotional \\
	\hline
	Reference & 3.26 \(\pm\) 0.24 & 3.39 \(\pm\) 0.25 \\
    Styletag & 3.21 \(\pm\) 0.22 & 3.66 \(\pm\) 0.21 \\
	\hline
	\hline
	\end{tabular}
\end{table}

\subsection{Case study}
We further investigated the possible application of our proposed model. By combining multiple style tags, our model enables detailed control of speaking style like \textit{angry and urgent}. By adding quantifiers to style tags such as \textit{little angry} or \textit{very angry}, it is possible to control the strength of expressed emotion. Also, we tested the generalization performance of our model with unseen style description. Even though most of the style tags in data are in the form of adjectives and adverbs, our model could also adapt to style description in sentences and nouns benefiting from the integration with the pretrained language model.

\section{Conclusion}
We proposed a novel method to control and transfer style with text description, which works just as well on a disjoint dataset. To learn multi-modal style embedding space, we integrated the pretrained language model. We adopted 3 style losses, style embedding loss, style reconstruction loss, and style contrastive loss to improve style controllability and cross-speaker style transfer. Experimental results show that the proposed approach achieved good performance on text-driven emotional style control and cross-speaker style transfer with disjoint dataset. 

\section{Acknowledgement}
This research is supported by Ministry of Culture, Sports and Tourism and Korea Creative Content
Agency(Project Number: R2021050007)

\newpage
\bibliographystyle{IEEEtran}
\bibliography{main}

\end{document}